\documentclass[pdflatex,sn-mathphys-num]{sn-jnl}

\usepackage{graphicx}%
\usepackage{multirow}%
\usepackage{amsmath,amssymb,amsfonts}%
\usepackage{amsthm}%
\usepackage{mathrsfs}%
\usepackage[title]{appendix}%
\usepackage{xcolor}%
\usepackage{textcomp}%
\usepackage{manyfoot}%
\usepackage{booktabs}%
\usepackage{algorithm}%
\usepackage{algorithmicx}%
\usepackage{algpseudocode}%
\usepackage{listings}%
\usepackage{tabularx}
\usepackage{array}
\usepackage{longtable}

\theoremstyle{thmstyleone}%
\theoremstyle{thmstyletwo}%

\theoremstyle{thmstylethree}%

\begin{document}

\title[Article Title]{BenCao: An Instruction-Tuned Large Language Model for Traditional Chinese Medicine}


\author[1,2]{\fnm{Jiacheng} \sur{Xie}}

\author[1,2]{\fnm{Yang } \sur{Yu}}

\author[3]{\fnm{Yibo} \sur{Chen}}

\author[1,2]{\fnm{Hanyao} \sur{Zhang}}

\author[4]{\fnm{Lening} \sur{Zhao}}

\author[5]{\fnm{Jiaxuan} \sur{He}}

\author[1,2]{\fnm{Lei} \sur{Jiang}}

\author[6]{\fnm{Xiaoting} \sur{Tang}}

\author[7]{\fnm{Guanghui} \sur{An}}

\author*[1,2]{\fnm{Dong} \sur{Xu}}

\affil[1]{\orgdiv{Department of Electrical Engineering and Computer Science}, \orgname{University of Missouri}, \orgaddress{ \city{Columbia},  \state{MO}, \country{USA}}}

\affil[2]{\orgdiv{Christopher S. Bond Life Sciences Center}, \orgname{University of Missouri}, \orgaddress{ \city{Columbia},  \state{MO}, \country{USA}}}

\affil[3]{\orgdiv{Institute for Data Science and Informatics}, \orgname{University of Missouri}, \orgaddress{ \city{Columbia},  \state{MO}, \country{USA}}}

\affil[4]{\orgdiv{School of Engineering and Applied Science}, \orgname{University of Pennsylvania}, \orgaddress{ \city{Philadelphia},  \state{PA}, \country{USA}}}

\affil[5]{\orgdiv{Department of Computer Science and Mathematics}, \orgname{Truman State University}, \orgaddress{ \city{Kirksville},  \state{MO}, \country{USA}}}

\affil[6]{\orgdiv{Community Health Service Center Shanghai Pudong New Area},  \orgaddress{ \city{Shanghai}, \country{China}}}

\affil[7]{\orgdiv{School of Acupuncture-Moxibustion and Tuina, Shanghai University of Traditional Chinese Medicine},  \orgaddress{ \city{Shanghai}, \country{China}}}


\abstract{Traditional Chinese Medicine (TCM), with a history spanning over two millennia, plays a role in global healthcare. However, applying large language models (LLMs) to TCM remains challenging due to its reliance on holistic reasoning, implicit logic, and multimodal diagnostic cues. Existing TCM-domain LLMs have made progress in text-based understanding but lack multimodal integration, interpretability, and clinical applicability. To address these limitations, we developed BenCao, a ChatGPT-based multimodal assistant for TCM integrating structured knowledge bases, diagnostic data, and expert feedback refinement. BenCao was trained through natural language instruction tuning rather than parameter retraining, aligning with expert-level reasoning and ethical norms specific to TCM. The system incorporates a comprehensive knowledge base of over 1,000 classical and modern texts, a scenario-based instruction framework for diverse interactions, a chain-of-thought simulation mechanism for interpretable reasoning, and a feedback refinement process involving licensed TCM practitioners. BenCao connects to external APIs for tongue-image classification and multimodal database retrieval, enabling dynamic access to diagnostic resources. In evaluations across single-choice question benchmarks and multimodal classification tasks, BenCao achieved superior accuracy to general-domain and TCM-domain models, particularly in diagnostics, herb recognition, and constitution classification. The model was deployed as an interactive application on the OpenAI GPTs Store, accessed by nearly 1,000 chats globally as of October 2025. This study demonstrates the feasibility of developing a TCM-domain LLM through natural language–based instruction tuning and multimodal integration, offering a practical framework for aligning generative AI with traditional medical reasoning and a scalable pathway for real-world deployment.}




\maketitle

\section{Introduction}\label{sec1}

Traditional Chinese Medicine (TCM) \cite{tang2008traditional}, with a history spanning more than two millennia, remains an important component of global health care. According to the World Health Organization\cite{world2002traditional}, more than one hundred countries report laws or regulatory frameworks related to traditional and complementary medicine, and many provide service coverage in routine health systems. In China alone, TCM institutions delivered approximately 1.28 billion clinical visits in 2023, underscoring the substantial scale of TCM-based prevention and care\cite{nmpa_tcm2023_2024}.

Parallel to this, the emergence of large language models (LLMs) such as GPT-4\cite{achiam2023gpt}, Gemini 2.5\cite{comanici2025gemini}, and Claude 3\cite{kurokawa2024diagnostic} has transformed the landscape of natural language understanding and reasoning. These models demonstrate remarkable capabilities in tasks involving knowledge retrieval, reasoning, and decision support, leading to wide applications across education\cite{wang2024large}, law\cite{lai2024large}, scientific research\cite{naveed2025comprehensive}, and healthcare\cite{singhal2025toward}. In clinical contexts, recent studies have shown that general-domain LLMs can assist in diagnosis generation, clinical documentation, and medical question answering, marking a significant step toward intelligent and data-driven healthcare systems\cite{thirunavukarasu2023large,singhal2023large,hager2024evaluation}.

Despite these advances, the application of LLMs within the TCM domain remains in its infancy. Unlike Western biomedicine, TCM emphasizes holistic reasoning and syndrome differentiation, which depend heavily on contextual interpretation, implicit logic, and multimodal cues\cite{tian2023can}. The complexity and symbolic nature of TCM knowledge pose unique challenges for language models, particularly in aligning linguistic patterns with clinical reasoning grounded in ancient textual corpora and empirical observations. Existing efforts to develop TCM-domain LLMs have primarily focused on textual data, while the integration of multimodal inputs such as tongue, pulse, and facial features remains limited. 

The absence of multimodal models in TCM represents a major challenge. The diagnostic framework of TCM, known as the Four Diagnostic Methods (inspection, listening and smelling, inquiry, and palpation), relies on integrating multiple sensory modalities, including visual, auditory, olfactory, linguistic, and tactile information. However, most existing research still focuses on single modality. Wei et al. introduced BianCang\cite{wei2025biancang}, a large language model specifically designed for TCM, which integrates real-world hospital records and high-quality corpora. By combining continuous pretraining with supervised fine-tuning, the model achieved notable improvements in TCM text comprehension and diagnostic reasoning tasks. The release of both model weights and datasets has further advanced systematic research in TCM language modeling. However, BianCang still exhibits limitations in clinical reasoning and generalization when confronted with complex consultation scenarios. To enhance diagnostic precision and interactive capability, Yan et al. subsequently proposed JingFang\cite{yang2025jingfang}, which incorporates a multi-agent collaborative reasoning framework and a syndrome restoration strategy. This design enables the model to emulate the reasoning processes observed in real-world clinical consultations, demonstrating a higher level of syndrome differentiation and treatment determination performance. Nevertheless, the model has not been publicly released, and its evaluation remains restricted mainly to symptom-based question answering, lacking comprehensive coverage of fundamental TCM theory and classical medical literature. Regarding openness and practical applicability, Dai et al. developed TCMChat\cite{dai2024tcmchat}, an open-source model instruction-tuned from Baichuan2-7B. They built datasets covering six major TCM application scenarios, including knowledge question answering, case reasoning, and prescription recommendation, and demonstrated improved performance over general-purpose language models across multiple benchmarks. The code and interfaces are publicly available on HuggingFace and GitHub, highlighting the model’s scalability and ease of use. However, TCMChat mainly addresses surface-level knowledge generation and lacks deeper clinical reasoning, interpretability, and multimodal capabilities such as processing tongue images or pulse signals, which are essential for comprehensive TCM diagnosis.

Although the TCM-domain LLMs have achieved notable progress at the algorithmic level, they still face substantial challenges in real-world deployment and clinical application. Most existing TCM models have released their weights or research-oriented demonstration versions, yet remain far from fully functional systems for clinical or public use. One major obstacle lies in their enormous parameter sizes and the resulting computational demands. Current TCM-domain LLMs comprise billions of parameters, requiring high-end GPUs or large-scale cluster computing for both training and inference, which leads to prohibitive deployment and maintenance costs beyond the reach of most healthcare institutions and independent developers.

Furthermore, these models exhibit limited accuracy and robustness in real medical tasks such as syndrome differentiation, prescription generation, and individualized diagnostic reasoning. While they can reproduce surface-level knowledge from classical TCM literature, their outputs frequently lack clinical reliability, logical coherence, and interpretability, which are fundamental for medical applications. As a result, a considerable gap persists between current research prototypes and deployable clinical systems. Achieving trustworthy, cost-efficient, and scalable deployment of TCM large language models, while ensuring safety, reliability, and controllability, remains a key challenge.

Building upon this motivation, we developed BenCao, a ChatGPT-based multimodal TCM assistant designed to bridge this gap. The system integrates textual and visual modalities and has been systematically evaluated against existing TCM-domain and general-domain LLMs across multiple real-world case studies.

The main contributions of this work are summarized as:

1.	Development of the first multimodal large language model for TCM. We present BenCao, a ChatGPT-based assistant capable of processing textual and visual information, including structured symptom descriptions, herb recognition, and tongue-image analysis, thereby enabling multimodal reasoning consistent with TCM diagnostic logic.

2.	Construction of a comprehensive domain-specific knowledge base and scenario-driven instruction framework. BenCao integrates over 1,000 classical and modern TCM texts and employs structured prompt design across four real-world interaction scenarios, allowing adaptive reasoning and dialogue behaviors aligned with authentic clinical and educational contexts.

3.	Implementation of human feedback–guided instruction refinement and chain-of-thought simulation. Through iterative expert interaction and interpretable reasoning design, BenCao aligns its responses with professional TCM logic, ethical safety constraints, and transparent stepwise reasoning patterns, enhancing reliability and interpretability without parameter retraining.

4.	Comprehensive evaluation across TCM and general-domain models. We conducted comparative evaluations of BenCao against both general-purpose and TCM-domain large language models across question-answering and visual recognition tasks, providing an in-depth analysis of model adaptability and reasoning performance in multimodal TCM applications.

\section{Methods}\label{sec2}

\subsection{Data Collection and Knowledge Base Construction}\label{subsec2}
We compiled a comprehensive TCM knowledge base consisting of over 1,000 classical texts and modern textbooks, providing the foundational corpus for BenCao’s domain understanding. The collected materials encompass several categories of data, including classical medical literature, standard teaching materials, clinical and experiential records, and multimodal resources such as diagnostic images. To enable BenCao to interpret and utilize the content effectively, all collected PDFs were converted into editable Word or TXT formats, followed by manual proofreading and correction to minimize information loss during OCR (Optical Character Recognition) processing. Irrelevant textual elements, such as prefaces, acknowledgments, or publication notes were removed during data-cleaning procedures to ensure content precision and relevance. We applied image compression and resolution optimization for high-resolution or large-scale medical illustrations to balance visual quality with computational efficiency. Given the file upload limitations of the knowledge interface, we merged related documents into consolidated files to reduce the total number of attachments while maintaining full content coverage. Subsequently, we linked BenCao to the curated knowledge base through structured prompt design, enabling the system to identify which reference files should be consulted for different categories of user queries. For example, when users inquire about fundamental TCM theories, BenCao is directed to consult Attachment 1: Huangdi Neijing; when questions involve tongue diagnosis, it references Attachment 2: Atlas of TCM Tongue Diagnosis. After establishing this knowledge alignment, we conducted iterative validation tests using representative samples from the evaluation set. BenCao’s responses were examined to ensure that the system accurately retrieved and integrated relevant information from the appropriate knowledge modules before deployment.

\subsection{Role Definition}\label{subsec2}
To ensure domain fidelity and professional interpretability, we initialized BenCao as a senior TCM physician persona with decades of clinical and teaching experience. This role definition is the foundation for personality alignment and domain-specific specialization during instruction design and fine-tuning. It is characterized by the following attributes: (1) mastery of fundamental TCM theories, including comprehensive understanding of Yin–Yang, Five Elements, Qi–Blood–Body Fluids, and Zang–Fu organ systems, as well as their dynamic physiological and pathological relationships; (2) familiarity with classical and modern canonical texts such as Huangdi Neijing, Shanghan Lun, Diagnostics of Traditional Chinese Medicine, and Formulas of Chinese Medicine; (3) extensive clinical experience in syndrome differentiation, treatment formulation, and integrative prescription adjustment across diverse clinical contexts; and (4) strong pedagogical and communication skills, enabling the model to articulate complex TCM concepts clearly for both medical trainees and general audiences. 

\subsection{Scenario Instruction Design}\label{subsec2}
 After completing the expert persona initialization, we further developed a scenario-based prompt system that aligns with real-world interaction needs in TCM, as shown in Table 1. BenCao is capable of dynamically adapting its dialogue style and knowledge retrieval strategy according to different user intents, thereby simulating authentic scenarios of teaching, clinical consultation, and health guidance. The system is structured around four fundamental scenarios: 
 
(1) Learning of TCM Theory. This scenario targets TCM learners and enthusiasts. The model explains fundamental theories and core concepts by citing authoritative classical sources. Each response must explicitly reference its knowledge source, such as specific books, textbooks, or credible online information, to enhance the reliability and traceability of the content. The language style balances academic rigor and accessibility, using metaphors, analogies, and illustrative examples to clarify complex or abstract TCM concepts and improve user comprehension. 

(2) Conditioning for Mild Health Discomforts. This scenario addresses users presenting mild and common discomforts such as headache, insomnia, poor appetite, or indigestion. The model applies TCM syndrome differentiation principles to preliminarily identify potential patterns (e.g., liver Qi stagnation, spleen-stomach deficiency, or heart-spleen insufficiency), and provides corresponding lifestyle, dietary, and rest recommendations. Two mandatory safety principles are enforced: (a) the safety disclaimer, which requires the model to remind users that its suggestions are for reference only and cannot replace clinical diagnosis or prescription; and (b) the risk advisory, which instructs the model to advise users to seek professional medical care if symptoms worsen or persist. 

(3) Constitution Assessment and Tongue Diagnosis. When users wish to understand their current health status or TCM constitution type, the model conducts an interactive questionnaire based on the Guidelines for Classification and Determination of TCM Constitution. Through guided questions about lifestyle habits, emotional states, and physical characteristics, BenCao assists users in self-assessment. Users may also upload tongue images, which the model analyzes using knowledge-base references on tongue features for integrated evaluation. The results are strictly preliminary and for reference only; BenCao must clearly remind users that such assessments do not substitute professional medical diagnosis and that further evaluation by a certified TCM practitioner is recommended if needed. 

(4) Daily Health Preservation and Seasonal Wellness Guidance. This scenario is designed for healthy individuals and sub-healthy populations. Drawing from the TCM theory of harmony between nature and human, the model provides personalized health-preserving advice in accordance with seasonal transitions. For example, it emphasizes nourishing the liver in spring, clearing the heart and relieving summer heat in summer, moistening the lungs and dispelling dryness in autumn, and tonifying the kidneys and consolidating essence in winter. BenCao integrates knowledge from canonical texts such as Huangdi Neijing and Treatise on Nourishing Life and Prolonging Longevity, offering evidence-based recommendations on seasonal diets, rest schedules, and daily wellness practices.

\begin{longtable}{@{}p{2cm}p{5cm}p{5cm}@{}}
\caption{Scenario-based Instruction Design for BenCao}\label{tab1}\\
\toprule
Scenario  & Example Prompt  & Constraint \\
\midrule
Learning of TCM Theory &
You are an experienced TCM scholar with a solid academic background and extensive teaching experience. When explaining TCM concepts, use precise yet accessible language to help learners understand classical theories. Your explanation should include: (1) a systematic interpretation of the theoretical origin of the TCM concept; (2) an explanation of its relevance to clinical practice; (3) citations from at least one authoritative source (e.g., Huangdi Neijing, Outline of TCM Theory); (4) maintenance of formal tone and academic clarity, avoiding excessive colloquial expressions. &
The content should demonstrate academic accuracy and linguistic consistency. References to authoritative sources (classics or textbooks) are required. The model must avoid offering medical or formula-based treatment advice. \\
Conditioning for Mild Health Discomforts & You are a licensed TCM physician. When users describe mild symptoms such as headache, insomnia, poor appetite, or indigestion, perform a preliminary analysis based on syndrome differentiation principles (e.g., liver Qi stagnation, spleen-stomach deficiency, heart-spleen insufficiency). Provide lifestyle, diet, and rest recommendations following the principle of “regulating the body through self-adjustment.” Ensure that: (1) all suggestions focus on gentle self-care and prevention; (2) each response explicitly includes safety and risk reminders; (3) if symptoms worsen or persist, advise users to seek formal medical attention. & Safety Constraint: Each response must include the disclaimer “The following content is for reference only and cannot replace professional diagnosis or prescription.” 
Ethical Disclaimer: If symptoms worsen or persist, users should be advised to visit a certified hospital or professional TCM physician promptly. Prescription-based or diagnostic statements are strictly prohibited.
 \\
Constitution Assessment and Tongue Diagnosis & You are a TCM specialist in constitution assessment and tongue diagnosis. When users wish to learn about their health condition or TCM constitution type, conduct an interactive dialogue to collect lifestyle, emotional, and physical information. Use the Guidelines for Classification and Determination of TCM Constitution (Chinese Association of TCM, 2009) for analysis. (1) Provide an explanation of constitution characteristics (e.g., Qi-deficient, Yin-deficient, phlegm-damp type); (2) describe tongue features (color, coating, shape, and moisture); (3) offer basic health-preservation suggestions on diet, exercise, and emotional regulation; (4) clarify that all outputs are for reference only and not diagnostic conclusions. & Safety Constraint: Each response must include the disclaimer “The following content is for reference only and cannot replace professional diagnosis or treatment.” Diagnostic or prescription outputs are prohibited. \\
Daily Health Preservation and Seasonal Wellness Guidance & You are a TCM expert in health preservation. Provide personalized health guidance based on the theory of “Harmony between Heaven and Human” (tian ren xiang ying) and “nourishment according to the four seasons.” (1) Explain how seasonal changes affect the body—for example, nourishing the liver in spring, clearing the heart in summer, moistening the lungs in autumn, and tonifying the kidneys in winter; (2) recommend 3–5 seasonal foods and their health benefits; (3) provide lifestyle and emotional suggestions following the principle of “nourishing in accordance with seasonal transitions.” & Safety Constraint: Each response must include the disclaimer “The following content is for reference only and cannot replace professional diagnosis or prescription.” The model must not recommend or generate herbal formulas. Focus only on general wellness and dietary advice. \\
\bottomrule
\end{longtable}

\subsection{Chain-of-Thought Simulation}\label{subsec2}
To emulate the reasoning patterns of experienced TCM clinicians in real-world diagnosis and teaching contexts, we incorporated a Chain-of-Thought (CoT) simulation mechanism into the BenCao system. This mechanism enables the model to exhibit a structured and interpretable reasoning process rather than providing direct, conclusion-oriented responses. When user-provided information is insufficient or ambiguous, the model is designed to proactively request additional input to prevent superficial or fragmented reasoning, thereby enhancing the reliability, personalization, and interpretability of its outputs. The CoT simulation follows the cognitive framework of TCM syndrome differentiation and treatment determination, organizing the reasoning process into four sequential stages: (1) Symptom Recognition: extracting information from the user’s chief complaints, accompanying symptoms, tongue features, and lifestyle patterns; (2) Pattern Differentiation: conducting diagnostic analysis according to TCM theoretical dimensions such as Yin–Yang, Five Elements, Zang–Fu, and Qi–Blood–Body Fluids; (3) Treatment Principle Reasoning: deriving therapeutic directions based on the identified syndrome pattern, such as soothing the liver and relieving stagnation, strengthening the spleen and tonifying Qi, or nourishing Yin and clearing heat; (4) Lifestyle Recommendation Generation: producing non-prescriptive suggestions on lifestyle and dietary adjustments, taking into account individual constitution and seasonal variations.

At each stage, the system performs an evidence sufficiency check to assess the completeness of key diagnostic factors, including cold–heat, deficiency–excess, interior–exterior, and Qi–Blood–Fluid attributes. If critical information is missing, the model enters an information-seeking phase, asking a limited number of high-information-gain questions (typically no more than three to five) to collect essential details for reasoning refinement. For instance, the model may inquire: “Do you often feel cold or have cold hands and feet?”, “Do you prefer cold drinks?”, “Is your stool currently dry or loose?”, or “Have you experienced significant mood changes recently?” to better infer the user’s cold–heat tendencies, Qi–Blood state, and emotional triggers.

This interactive reasoning loop continues until one of three termination conditions is reached: (a) the coverage of core diagnostic elements exceeds 80\%, indicating sufficient information for reliable reasoning; (b) the incremental information gain over two consecutive inquiry rounds falls below 10\%, suggesting that further questioning would yield minimal benefit; or (c) the user explicitly declines to provide additional information. Once these conditions are met, BenCao automatically transitions into a conservative compliant mode, offering only preliminary constitution analysis and general lifestyle recommendations while strictly avoiding any diagnostic or prescription-related outputs. In scenarios involving high uncertainty or ethically sensitive contexts, such as pregnancy, pediatric cases, chronic diseases, or acute severe symptoms, the system activates an uncertainty-aware safeguard mechanism. In such cases, BenCao generates probabilistic and cautious statements, accompanied by explicit disclaimers and medical advisories, reminding users that if symptoms worsen, they should promptly seek professional evaluation at a certified medical institution.

\subsection{Human Feedback–Guided Instruction Refinement}\label{subsec2}
In this stage, we implemented a human feedback-guided instruction refinement mechanism to adjust BenCao’s behavior through iterative expert interaction. Unlike conventional approaches that rely on parameter retraining, this refinement process focused on semantic-level optimization, using expert feedback in natural language to gradually shape the model’s reasoning structure, linguistic precision, and ethical compliance. The objective was to align BenCao’s responses with the logic, terminology, and professional standards of TCM. We evaluated the model across the four core interactive scenarios and analyzed its generated outputs. When the responses failed to meet the established requirements, for example, lacking citations from authoritative sources, presenting incomplete syndrome differentiation logic, or omitting safety reminders evaluators provided critical feedback in natural language, explicitly identifying the errors and suggesting targeted improvements. Conversely, when the model’s outputs met expectations regarding content accuracy, reasoning completeness, and ethical consistency, evaluators provided positive feedback, acknowledging strengths and reinforcing desirable behaviors. To ensure feedback quality and domain relevance, three licensed TCM physicians participated in the iterative refinement process. Each expert had more than ten years of clinical experience and specialized expertise in TCM diagnostics, prescription formulation, and constitutional theory. They served dual roles: as practitioners, they assessed the practicality and reliability in simulated clinical consultation and health education; as reviewers, they continuously refined the model’s behavioral boundaries and knowledge precision through cycles of critical and positive feedback. This feedback-driven optimization enabled BenCao to progressively acquire a reasoning style and communication pattern consistent with expert-level TCM practice, enhancing its factual accuracy, interpretability, and adherence to ethical standards.

\subsection{External API Integration}\label{subsec2}
To further enhance the knowledge coverage and reasoning reliability in complex TCM tasks, we integrated the system with previously developed tongue image classification models\cite{xie2021digital} and a multimodal TCM knowledge database\cite{xie2025tcm}. This integration was achieved through standardized API interfaces, enabling BenCao to dynamically connect with external models and structured data resources during response generation. Through this mechanism, BenCao can retrieve, reference, and synthesize multimodal information—such as image-derived diagnostic features, textual knowledge entries, and structured case data—thereby improving its interpretability, factual grounding, and consistency in handling real-world clinical reasoning scenarios.

\section{Results}\label{sec3}
\subsection{Performance across single-choice question categories in TCM}\label{subsec3}

To comprehensively assess the reasoning and domain adaptability of LLMs in TCM, we evaluated their accuracy on single-choice questions spanning seven representative TCM disciplines, including diagnostics, pharmacognosy, surgery, herbal formulas, and internal medicine as shown in Figure 1. Overall, general-domain models such as GPT-4o, Gemini 2.5 Pro, Grok3, and Claude 3 demonstrated varying accuracy across categories, with GPT-4o and Gemini 2.5 Pro generally performing more strongly. Notably, BenCao consistently achieved competitive or superior accuracy across multiple disciplines, particularly in diagnostics and herbal formulas—reflecting its enhanced capacity to capture domain-specific reasoning patterns characteristic of TCM. These findings suggest that BenCao possesses an improved understanding of both theoretical knowledge and clinical reasoning, underscoring its potential as a domain-adapted LLM for TCM education and decision support.

\begin{figure}[h]
\centering
\includegraphics[width=0.9\textwidth]{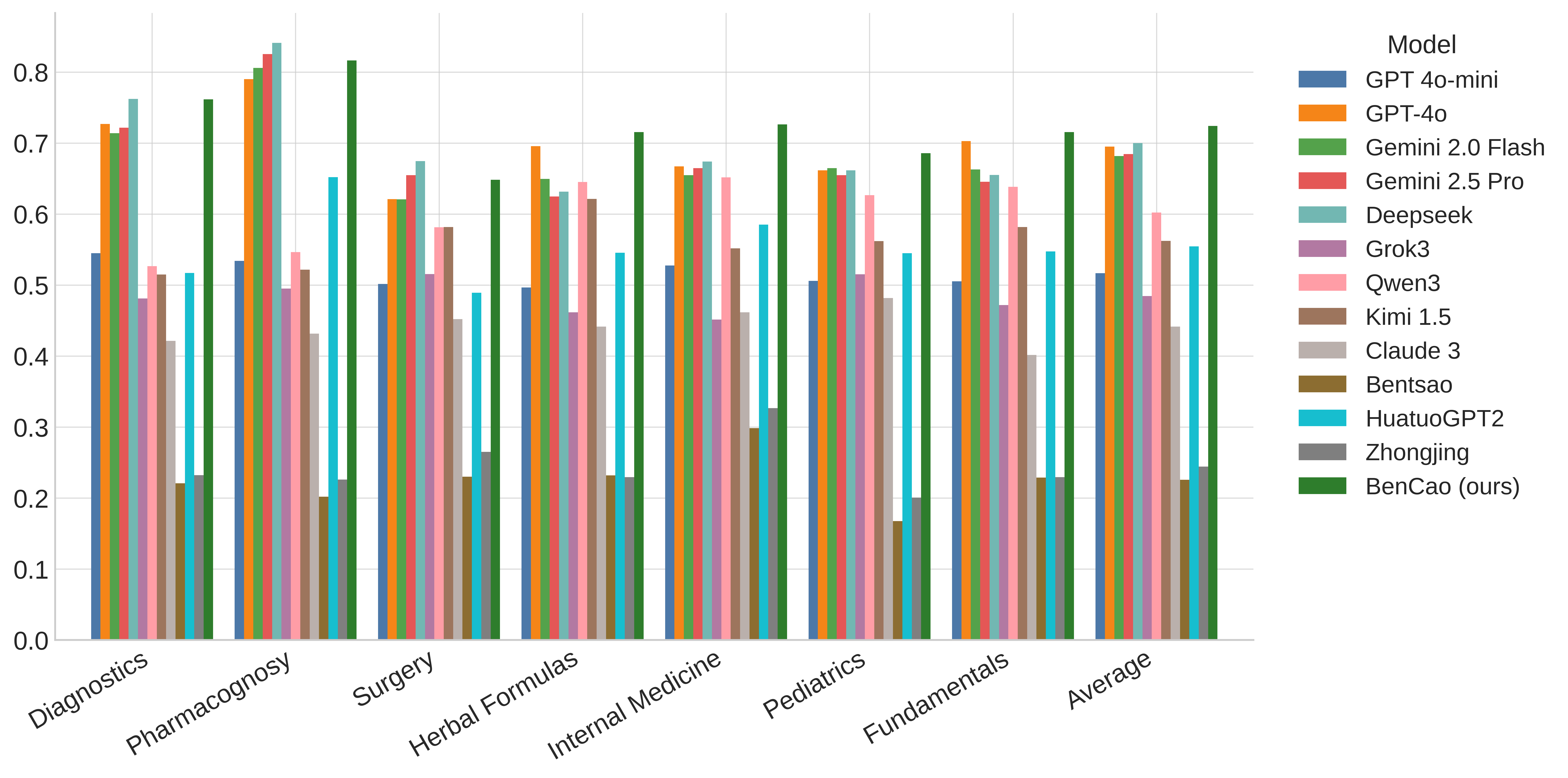}
\caption{Model performance across single-choice question categories in TCM. Accuracy of different LLMs on single-choice questions across seven TCM categories, including diagnostics, pharmacognosy, surgery, herbal formulas, and internal medicine. The x-axis indicates the question categories, and the y-axis shows the corresponding accuracy achieved by each model.}\label{fig1}
\end{figure}

\subsection{Performance across herb recognition and TCM constitution classification}\label{subsec3}

We compared the performance of LLMs on two TCM tasks: herb recognition and TCM constitution classification as shown in Figure 2. Across both tasks, general-domain LLMs exhibited notable variability in accuracy. BenCao achieved the highest performance, reaching 82.18 \% on herb recognition and 63.42 \% on constitution classification. Among the general-purpose models, Gemini 2.5 Pro showed the strongest performance on herb recognition (77.78 \%), while Qwen3, GPT-4o, and Gemini 2.5 Pro achieved moderate accuracy on constitution classification (57.86 \%, 52.90 \%, and 54.15 \%, respectively). Models such as Kimi 1.5 and Grok3 showed relatively low scores in both tasks. These results demonstrate that BenCao consistently outperforms general-domain LLMs across different TCM task types.

\begin{figure}[h]
\centering
\includegraphics[width=0.75\textwidth]{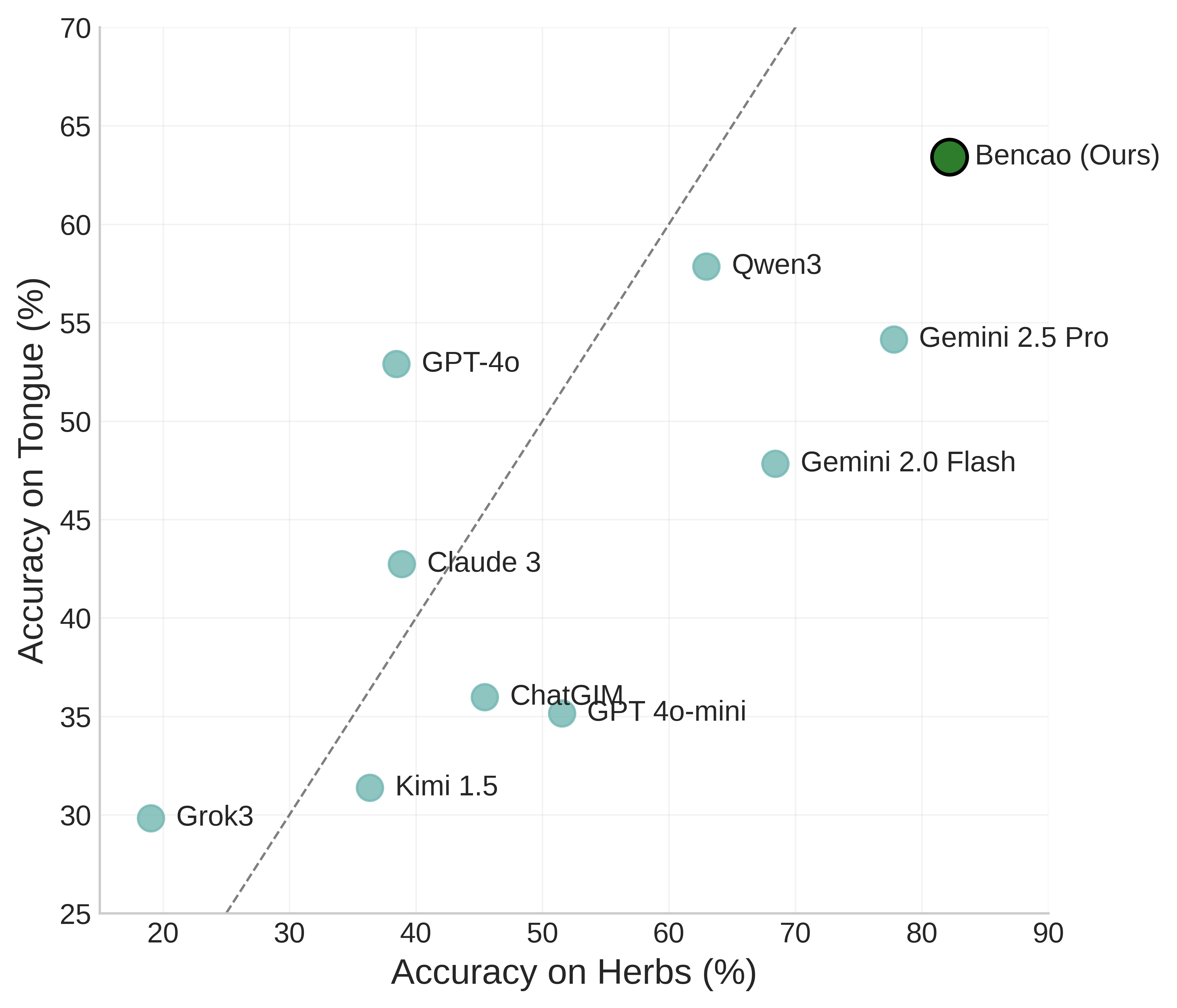}
\caption{Cross-task performance on the herb recognition and TCM constitution classification tasks. Each point denotes an individual LLM’s accuracy on two TCM tasks: herb recognition (x-axis) and TCM constitution classification (y-axis). The dashed diagonal indicates equal performance on both tasks; points toward the upper-right reflect stronger cross-task performance.}\label{fig2}
\end{figure}

\subsection{Deployment in the GPTs Store}\label{subsec3}
Following system development and validation, BenCao was deployed as an interactive intelligent agent on the OpenAI GPTs Store, enabling global users to access the TCM dialogue system through natural language interaction. Users can directly engage with the model without additional software installation or specialized medical expertise. By searching for “BenCao” within the GPTs Store, they can initiate real-time conversations and explore the system’s full range of capabilities. As of October 2025, BenCao has facilitated nearly 1,000 user interaction sessions. 

\section{Discussion}\label{sec4}
This study introduces BenCao, a ChatGPT-based multimodal assistant for TCM. BenCao demonstrates that alignment between expert reasoning and language models can be achieved without retraining model parameters by integrating structured knowledge bases, multimodal diagnostic resources, and human feedback-guided instruction refinement. The system performs effectively across a variety of TCM tasks, including theoretical explanation, case-based reasoning, constitution analysis, and wellness guidance, highlighting its potential as an intelligent platform for medical education, public health communication, and preliminary clinical support.

BenCao advances the current development of TCM-domain LLMs in several aspects. First, its knowledge base integrates more than one thousand classical and modern TCM sources, achieving both historical comprehensiveness and contemporary relevance. Second, its scenario-based instruction framework enables the model to adapt dialogue strategies to different user intents, better reflecting authentic interactions between physicians and patients or teachers and students. Third, the chain-of-thought simulation allows BenCao to display transparent and stepwise reasoning processes that mirror the cognitive framework of syndrome differentiation and treatment determination in TCM. Finally, the human feedback-guided instruction refinement mechanism ensures that the model’s outputs conform to professional logic, linguistic precision, and ethical norms while maintaining interpretability and user safety. Together, these components illustrate a practical approach to achieving human-centered alignment of LLMs in non-Western medical systems.

Despite these advances, BenCao remains a research-oriented prototype rather than a clinical decision-support system. Although the model performs well in structured question answering and reasoning tasks, its diagnostic accuracy and prescription-related capabilities are intentionally limited to prevent clinical misuse. The system’s reliance on textual data and a few visual modalities also constrains its applicability to more complex diagnostic contexts, such as pulse analysis or integrative treatment planning. Moreover, while human feedback refinement improved the model’s behavioral stability and adherence to ethical guidelines, it cannot yet replace the rigor of reinforcement learning or comprehensive benchmark evaluation.
Future work will focus on expanding BenCao’s multimodal integration and real-world validation. Incorporating physiological signals, electronic medical records, and multi-institutional expert annotations will allow for more comprehensive TCM diagnostic and therapeutic reasoning modeling. The creation of an open benchmark for TCM-domain LLMs with standardized evaluation protocols and ethical frameworks will be essential for ensuring safety, transparency, and reproducibility. In the long term, combining expert feedback with reinforcement learning based on clinical data may lead to adaptive and continuously improving TCM agents capable of supporting education, research, and patient engagement.

In summary, the development of BenCao represents an important step toward building trustworthy, knowledge-grounded, and human-aligned large language models for Traditional Chinese Medicine. By integrating multimodal reasoning, expert feedback, and safety mechanisms, this study demonstrates the potential of generative artificial intelligence to promote medical diversity and drive global healthcare innovation. It also provides a practical and scalable pathway for lightweight deploying TCM models and other domain-specific applications in real-world settings.

\section{Conclusion}\label{sec5}

This work presents BenCao, a ChatGPT-based multimodal assistant for TCM. By integrating structured knowledge, multimodal reasoning, and expert feedback, BenCao achieves reliable, interpretable, and context-aware interactions that align with the diagnostic and reasoning framework of TCM. The system demonstrates strong potential in educational, consultative, and health-guidance contexts, offering an early example of how large language models can be adapted for culturally grounded and expert-driven medical domains. Furthermore, this work demonstrates the feasibility of developing a TCM-domain large language model through natural language–based instruction tuning, without the need for parameter retraining. The approach provides a practical and generalizable reference framework for the real-world adaptation of other foundation models in specialized domains.

\bibliography{BenCao}

\end{document}